\documentclass{article}
\usepackage{lineno}
\modulolinenumbers[5]
\usepackage{amssymb}
\usepackage{bm}
\usepackage{float}
\usepackage{wrapfig}
\usepackage{geometry}
\usepackage{caption}
\usepackage{relsize}

\usepackage{url}            
\usepackage{booktabs}       
\usepackage{amsfonts}       
\usepackage{nicefrac}       
\usepackage{microtype}      
\usepackage{amsmath}
\usepackage{subcaption}
\usepackage{algorithm}
\usepackage[noend]{algpseudocode}

\usepackage{graphicx}
\usepackage{float}
\usepackage{caption}
\usepackage{graphicx}
\usepackage{wrapfig}

\newcommand\tab[1][1cm]{\hspace*{#1}}
\usepackage{bm}
\newcommand{\bx}{\mathrm{\mathbf{x}}}
\newcommand{\bcx}{\mathrm{\mathbf{X}}}
\newcommand{\bca}{\mathrm{\mathbf{A}}}
\newcommand{\bcs}{\mathrm{\mathbf{S}}}

\newcommand{\bce}{\mathrm{\mathbf{E}}}
\newcommand{\be}{\mathrm{\mathbf{e}}}

\begin{document}

\begin{center}
		\large{ \textbf{Large-scale nonlinear Granger causality: A data-driven, multivariate approach to recovering directed networks from short time-series data } }\\
	
\end{center}



Axel Wism\"uller$^{1,2,3,4,a}$, Adora M. DSouza $^{1,a,*}$, Anas Z. Abidin$^{2}$ \\

$^1$Department of Electrical Engineering, University of Rochester, Rochester, New York, USA
	
$^2$Department of Biomedical Engineering, University of Rochester, Rochester, New York, USA

$^3$Department of Imaging Sciences, University of Rochester, NY, USA

$^4$Faculty of Medicine \& Institute of Clinical Radiology, Ludwig Maximilian Univ, Munich,Germany

$^a$Both authors contributed equally

$^*$Corresponding author (adora.dsouza@rochester.edu)

\vspace*{0.4in}

\begin{abstract}

To gain insight into complex systems it is a key challenge to infer nonlinear causal directional relations from observational time-series data. Specifically, estimating causal relationships between interacting components in large systems with only short recordings over few temporal observations remains an important, yet unresolved problem. Here, we introduce a large-scale Nonlinear Granger Causality (lsNGC) approach for inferring directional, nonlinear, multivariate causal interactions between system components from short high-dimensional time-series recordings. By modeling interactions with nonlinear state-space transformations from limited observational data, lsNGC identifies casual relations with no explicit a priori assumptions on functional interdependence between component time-series in a computationally efficient manner. Additionally, our method provides a mathematical formulation revealing statistical significance of inferred causal relations. We extensively study the ability of lsNGC to recovering network structure from two-node to thirty-four node chaotic time-series systems. Our results suggest that lsNGC captures meaningful interactions from limited observational data, where it performs favorably when compared to traditionally used methods. Finally, we demonstrate the applicability of lsNGC to estimating causality in large, real-world systems by inferring directional nonlinear, multivariate causal relationships among a large number of relatively short time-series acquired from functional Magnetic Resonance Imaging (fMRI) data of the human brain.

\end{abstract}




\section{Introduction}

Detecting causal influences between components of a complex system, especially from simultaneously observed time-series, is an actively growing area of research \cite{lacasa2015network, gao2017complex, dsouza2017exploring, dsouza2018mutual}. The cause-effect relationships quantified by any approach can be represented as a network graph. Networks are ubiquitous in natural as well as man-made systems. Examples of networks are interactions between individual neurons, regions in the brain, protein interaction, genetic networks, etc. These network graphs can be analyzed to reveal important properties of the system being studied. For example, investigating such a graph constructed from brain activity of healthy subjects and patients with some form of neurodegeneration can reveal vital information useful for diagnosis and treatment.

One of the most widely used approaches for estimating causal relations from time-series data is Granger causality analysis \cite{Granger1969}. It estimates causal influence from one time-series to another, if the prediction quality of the influenced time-series improves when the past of an influencer time-series is used, as compared to its prediction quality when the past of the influencer is not used. The definition of Granger causality (GC) establishes causal interactions by having a strict flow of time. Since it was originally formulated for linear models, its application to nonlinear formulations was treated with skepticism. However the problem of GC was extended to non-linear systems with promising results \cite{chen2004analyzing}. Ideally, a causality analysis method should 1) be multivariate, 2) be able to capture nonlinear dependencies, 3) work for systems with large number of variables, and 4) be data-driven. While GC is a multivariate analysis approach with both linear and nonlinear variants, it, very often, cannot be extended to large systems since the problem becomes an ill-posed, underdetermined one.

\label{multi_need}
In systems containing more than two time-series, performing a bivariate analysis - i.e. considering only pairs of time-series at a time without taking into account the effects of confounding variables - results in spurious causalities as a consequence of indirect connections \cite{Stephan2010}. In \cite{Blinowska2004}, the authors have shown that a bivariate analysis could result in misleading information regarding propagation of influence, while multivariate analysis distinguishes direct from indirect influences. Although a multivariate analysis can produce better estimates \cite{Geweke1984} of casual relations, the complexity of the model increases with increasing number of time-series in the system, hence making it computationally infeasible. Additionally, redundant variables can lead to underestimation of causal influences \cite{angelini2010redundant}. In summary, an approach that estimates multivariate interactions, while reducing redundancy, and being computationally feasible would be desired. 

Furthermore, most systems in nature show complex dynamics which cannot be represented well by linear approaches \cite{sugihara2012detecting,bischi2010nonlinear,sugihara1998nonlinear}. Nonlinear time-series analysis approaches have the advantage of capturing interaction patterns among components of such systems. Nonlinear models are more realistic, however, at the cost of increased computation time, and possible increase in number of parameters to be estimated. For linear stochastic systems, using nonlinear analysis approaches may not provide any benefit and could produce worse results. Nevertheless, ideally, an approach that can possibly capture multivariate, nonlinear interactions in large systems is desired. 

Non-linear extensions of GC have been proposed in recent literature \cite{chen2004analyzing}, including, kernel-based non-linear GC approaches \cite{liao2009kernel, marinazzo2008kernel, marinazzo2011nonlinear} has gained most traction. However, nonlinear Granger causality methods, such as those cited above, have not been applied to large systems. Possible reasons for this restrictions are: Besides computational expense, the extendibility to multivariate analysis of high-dimensional dynamic systems based on a low number of temporal observations, involves specific challenges regarding parameter optimization of sophisticated nonlinear time-series models on limited data. In this work, we address this bottleneck and introduce a large-scale Non-Linear Granger Causality (lsNGC) approach for estimating network structure. By introducing a nonlinear dimension reduction step, lsNGC aims at directed, multivariate time-series causality analysis in large complex networks. We demonstrate the applicability of lsNGC on estimating connectivity from resting-state functional MRI. However, lsNGC may be useful for other domains as well, given that the data is represented as signals simultaneously acquired. 

In the following sections we discuss lsNGC and the various networks we investigate. We compare our approach with two standard methods: 1) Kernel Granger causality \cite{marinazzo2008kernel}, which is multivariate and nonlinear, and 2) an approach that is bivariate and nonlinear, using local models (LM) \cite{sugihara2012detecting} to estimate causal influences. We test performance of simulated data with known ground truth of connections. Additionally, we demonstrate the application of the proposed lsNGC approach on real time-series data recorded using functional Magnetic Resonance Imaging (fMRI) from subjects presenting with symptoms of HIV associated neurocognitive disorder (HAND) and healthy controls. If lsNGC measures can characterize brain connectivity well, it should be useful in distinguishing the two subject groups.

\section{Methods}

\subsection{Large-scale Nonlinear Granger causality}
\label{lsNGCexp}
Large-scale nonlinear Granger causality adopts theoretical concepts from Granger causality analysis. Granger causality (GC) is based on the concept of time-series precedence and predictability; here, the improvement in the prediction quality of a time-series in the presence of the past of another time-series is quantified. This reveals if the predicted time-series was influenced by the time-series whose past was used in the prediction, uncovering the causal relationship between the two series \cite{Granger1969} under investigation. The supplementary material (section 1) details the theoretical concepts involved in Granger causality analysis.

LsNGC estimates causal relationships by first creating a nonlinear transformation of the state-space representation of the time-series, whose influence on others is to be measured, and another representation of the rest of the time-series in the system. Consider a system with $N$ time-series, each with $T$ temporal samples. Let the time-series ensemble \- $\bcx \in \mathbb{R}^{N \times T}$ be 
${\bcx = (\bx_1, \bx_2, ..., \bx_N)^T}$, where $\bx_n \in \mathbb{R}^{T}$, $ n \in \{1, 2, ..., N\}$, $\bx_n = (x_n(1), x_n(2), ..., x_n(T))$. The time-series ensemble $\bcx$ can also be represented as $\bcx = (\bx(1), \bx(2), ..., \bx(T))$, where $\bx(t) \in \mathbb{R}^{N \times 1}$, \-$t \in \{1, 2, ..., T\}, \bx(t) = (x_1(t), x_2(t), x_3(t), ..., x_N (t))^T$. Let's say that we are interested to learn if $\bx_s$ influences $\bx_r$. We first construct the phase space representation of $\bx_s$ with embedding dimension $d$, as $\bm{W}_s$. The state at time $t$ is $\bm{w}_s(t) = [x_s(t-(d-1)), ~...,~ x_s(t-1),~ x_s(t)]$, and $t \in {d, ..., T-1}$. Say we are interested in quantifying the influence of $\bx_s$ on $\bx_r$ in the presence of all confounding variables and also by modeling nonlinearities present in the data. Confounding variables can be accounted for by performing a multivariate analysis. Additionally, we account for nonlinear interactions among time-series by transforming the original space using a nonlinear transformation function. 

To perform a multivariate analysis, it is desirable to have a phase space reconstruction, where prediction is performed using all the time-series, apart from $\bx_s$ whose influence is to be quantified. From the time-series ensemble $\bcx\backslash\bx_s$ we construct the phase space reconstruction $\bm{Z}_s$. The state of this multivariate system at a given time-point is 

\[
\bm{z}_s(t) =\left\{
\begin{array}{ll}
x_1(t-(d-1)), ..., x_1(t-1), x_1(t), ...\\
x_2(t-(d-1)), ..., x_2(t-1), x_2(t), ...\\	
\tab .\\
\tab .\\
x_{N-1}(t-(d-1)), ..., x_{N-1}(t-1), x_{N-1}(t)\\
\end{array}
\right.
\]
It should be noted that $\bm{Z}_s$ does not contain any terms from $\bx_s$. 

In brief, we have constructed two systems represented by phase states $\bm{W}_s$ and $\bm{Z}_s$. $\bm{W}_s$ represents the states of only the time-series whose influence we want to quantify, i.e $\bx_s$, and $\bm{Z}_s$ represents the multivariate phase space incorporating all time-series but $\bx_s$.

Coming back to Granger causality (GC), GC works on the principle that if the prediction quality of a time-series $\bx_r$ improves in the presence of $\bx_s$ as compared to its prediction quality in the absence of $\bx_s$, having considered the rest of the time-series in both models, then $\bx_s$ Granger causes $\bx_r$. It quantifies boost in the prediction quality, by comparing two models, one that uses information from the states of $\bx_s$ and the other that does not. Let $\mathbf{f}$ and $\mathbf{g}$ represent two nonlinear functions. The two estimates of $\bx_r$ are given by:

\begin{equation}
\label{eq:unrestrct1}
\hat{\mathbf{x}}_{r, s}=\mathbf{a}_{11}\mathbf{f}(\bm{Z}_s) + \mathbf{a}_{12}\mathbf{g}(\bm{W}_s)
\end{equation}
\begin{equation}
\label{eq:restrct1}
{\tilde{\mathbf{x}}}_{r,s}=\mathbf{b}_{1}\mathbf{f}(\bm{Z}_s) 
\end{equation}

In the above equations, $\mathbf{a}$ and $\mathbf{b}$ are the weights or model parameters, obtained by minimizing the mean squared errors in the estimate of $\bx_r$. The quantities $\hat{\mathbf{x}}_{r, s}$ and ${\tilde{\mathbf{x}}}_{r,s}$ are the estimates of $\bx_r$ calculated by the two models. The subscript $(r, s)$ denotes that these models were constructed to investigate the influence of $\bx_s$ on $\bx_r$. 
In this study, we use the generalized radial basis function (GRBF) as nonlinear transformations $\mathbf{f}$ and $\mathbf{g}$. Mathematical formulation of the GRBF is provided in the supplementary material (section 2.2). In brief, representative clusters of the state space $\bm{Z}_s$ and $\bm{W}_s$ are obtained using clustering methods, such as $k$-means clustering, where $k$ can be seen as the number of hidden neurons in a GRBF neural network. Let $c_z$ and $c_w$ be the number of hidden layer neurons for $\bm{Z}_s$ and $\bm{W}_s$, respectively. 

The $f$-statistic can be obtained by recognizing that the two models, equations (\ref{eq:unrestrct1}) and (\ref{eq:restrct1}), can be characterized as the unrestricted model and the restricted model, respectively. Residual sum of squares ($RSS$) of the restricted model, $RSS_{R}$, and residual sum of squares of the unrestricted model, $RSS_{U}$, are obtained. Here, $n$ is the number of time-delayed vectors, $p_{U}$ and $p_{R}$ are the number of parameters to be estimated for the unrestricted and restricted model, respectively. For equations (\ref{eq:unrestrct1}) and (\ref{eq:restrct1}), $p_{U} = c_z + c_w$ and $p_{R} = c_z$, respectively. 



A measure of lsNGC can be obtained using the $f$-statistic, given by:
\begin{equation}
\label{eq:ftest}
F_{\bx_s\to\bx_r}  = \frac{(RSS_{R} - RSS_{U})/(p_{U} - p_{R})}{(RSS_{U})/(n - p_{U} - 1)}
\end{equation}

$F_{\bx_s\to\bx_r}$ quantifies the influence of $\bx_s$ on $\bx_r$, by testing the equality of variances of errors in prediction of the $\bx_r$ by both the models i.e. equations (\ref{eq:unrestrct1}) and (\ref{eq:restrct1}). If the variance of the error in predicting $\bx_r$ is lower when $\bx_s$  is used, then $\bx_s$ is said to Granger cause $\bx_r$. The measure $F_{\bx_s\to\bx_r}$ is stored in the affinity matrix $\bcs$ at position $(\bcs)_{s,r}$, where $\bcs$ is an $N \times N$ matrix of lsNGC indices. Each lsNGC measure in the affinity matrix can be represented as a directed edge connecting the $s^{\mathrm{th}}$ node to the $r^{\mathrm{th}}$ node in a network graph. Implementation specifics that make lsNGC computationally efficient and various parameter information are provided in the supplementary material, section 2. We will also make publicly available the lsNGC MATLAB software.

\section{Results}

To evaluate the approach, several benchmark simulations are considered and performance is compared to two state of the art approaches, Kernel Granger Causality (KGC) \cite{marinazzo2008kernel} and mutual nonlinear cross-mapping methods \cite{sugihara2012detecting} using local models (LM). These two approaches are discussed briefly in the supplementary material, section 5.

\begin{figure}[h]
	\centering
	\includegraphics[scale=0.6]{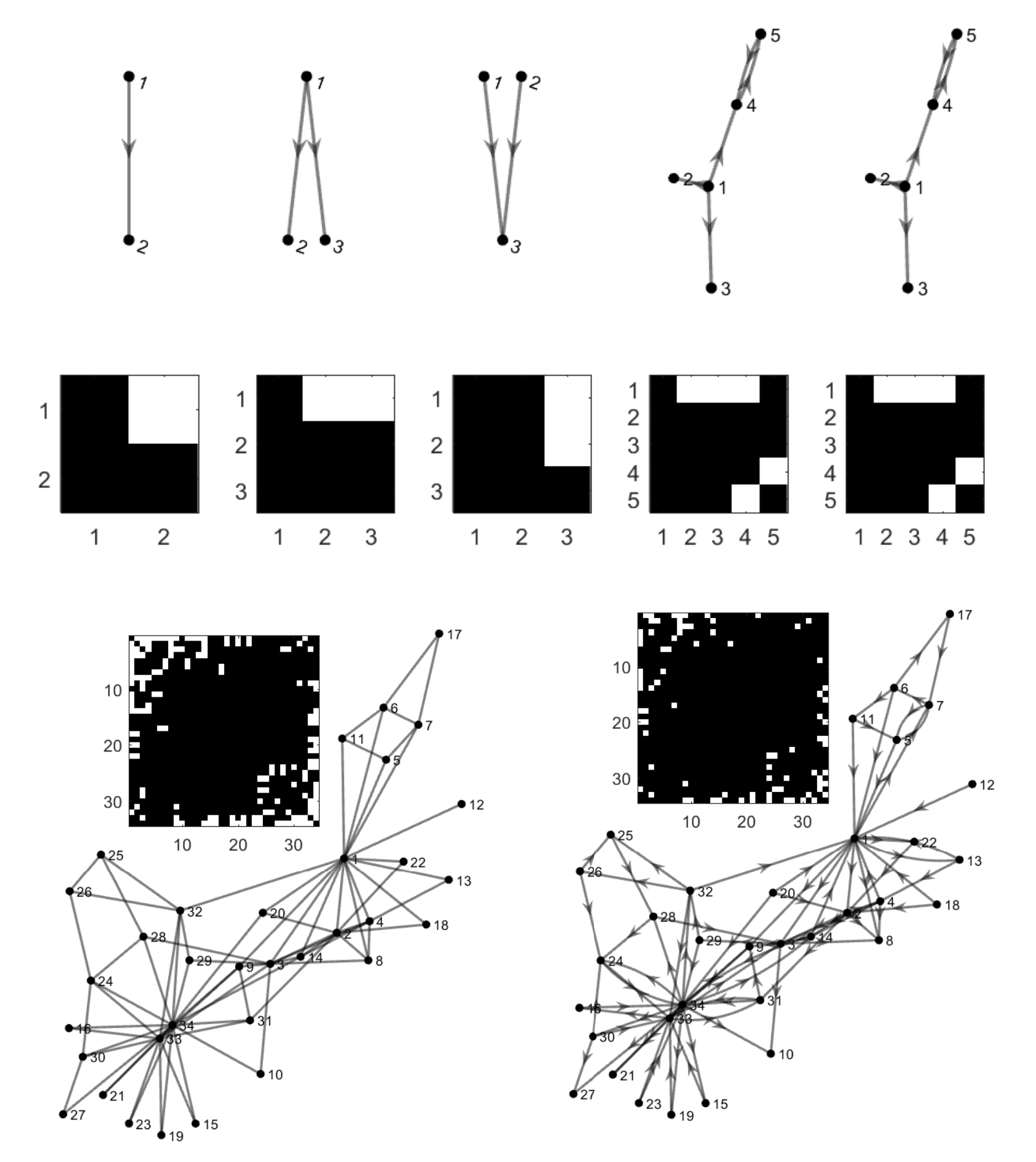}
	\caption{Different network structures and their corresponding adjacency matrices. Going from top left to bottom right from the first column to the seventh, we have the 2-species, 3-fan out, 3-fan in, 5-linear, 5-nonlinear, 34-Zachary1 and 34-Zachary2 networks.}
	\label{fig:networks}
\end{figure}

\subsection*{Simulated data network models}

We begin by creating benchmark simulations. Fifty different sets of each type of simulation were created, this is useful for estimating the consistency of the method. All the simulations were generated to have 500 time-points. 

\paragraph*{Two species logistic model:}

\begin{figure}[h]
	\centering
	\includegraphics[scale=0.7]{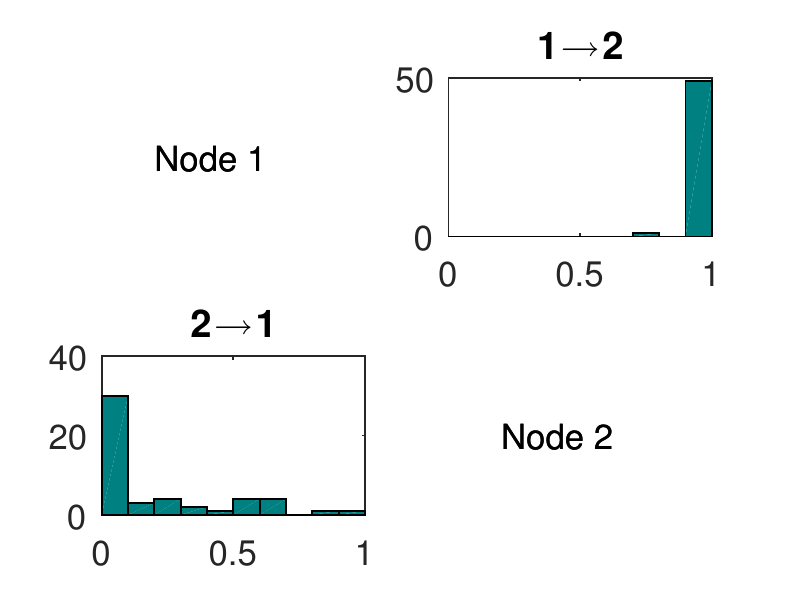}
	\caption{Histogram of scores (normalized between 0 to 1) obtained by lsNGC for the 2-logistic network over 50 different sets of the simulation. The influence of $x_1$ on $x_2$ is captured quite well across the 50 simulations.}
	\label{fig:hist_2_logistic}
\end{figure}

\label{two_species}
Before investigating empirical data or systems with a large number of time-series, it is imperative to test performance on a simple network structure with directed interaction. To this end, the two species logistic model which is one of the commonly studied \cite{ma2014detecting} chaotic time-series systems is considered:

\begin{equation}
\begin{gathered}
x_1(t+1)=x_1(t)[r_1-r_1x_1(t)-\gamma_{1,2}x_2(t)] \\
x_2(t+1)=x_2(t)[r_2-r_2x_2(t)-\gamma_{2,1}x_1(t)]
\end{gathered}
\end{equation}
\noindent 
where $r_1 = 3.7$, $r_2 = 3.8$, and $\gamma_{1,2}$ and $\gamma_{2,1}$  are the coupling constants. We adopt all values used from \cite{sugihara2012detecting,ma2014detecting}. For the unidirectional case, the coupling constants take the values $\gamma_{2,1} = 0.32$, and $\gamma_{1,2} = 0$. Uniformly distributed random numbers between [0, 1] are used as initial conditions and the first 50 time points are discarded. In our results, we refer to this network as 2-logistic (Figure \ref{fig:networks}). All the lsNGC scores are estimated using \ref{eq:ftest}. Figure \ref{fig:hist_2_logistic} is a histogram of the lsNGC scores (normalized between 0 and 1 using min-max normalization for display), assigned to $x_1 \to x_2$ and $x_2 \to x_1$ over the 50 different sets of the simulation. LsNGC is able to capture directed connections from $x_1 \to x_2$ well which is evident from Figure \ref{fig:hist_2_logistic} by the high scores assigned to $x_1 \to x_2$ compared to $x_2 \to x_1$. Detailed comparative quantitative results for the performance of various algorithms on all simulated networks shown in Figure \ref{fig:networks} are presented in Figures \ref{fig:AUC_res}, \ref{fig:sens_spec}, and \ref{fig:time_lengths}. 

\paragraph{Complex system with three nodes:}
Following this, we consider a slightly more complicated system with three nodes. The coupling parameters of this system can be set such that it can show fan-in and fan-out causality, as can be seen in Figure \ref{fig:networks}. 

\begin{equation}
x_j(t+1)=x_j(t)\Big(\gamma_{jj} - \sum_{i=1,2,3}^{}{\gamma_{ji}x_i(t)}\Big), j = 1,2,3
\end{equation}

Here $\gamma_{ji}$ are the coupling parameters. Again, we adopt parameter values from \cite{ma2014detecting}. In the fan-out case, $\gamma_{11} = 4$, $\gamma_{22} = 3.1$, $\gamma_{33} = 2.12$, $\gamma_{21} = 0.21$ and $\gamma_{31} = -0.636$, the other parameters are zero (3-fan out). In the fan-in case, $\gamma_{11} = 4$, $\gamma_{22} = 3.6$, $\gamma_{33} = 2.12$, $\gamma_{31} = 0.636$ and $\gamma_{32} = -0.636$, the other parameters are zero  (3-fan in). Uniformly distributed random numbers between [0, 1] are used as initial conditions and the first 50 time points are discarded.

\begin{figure}[h]
	\centering
	\begin{subfigure}{.5\textwidth}
		\centering
		\includegraphics[width=1\linewidth]{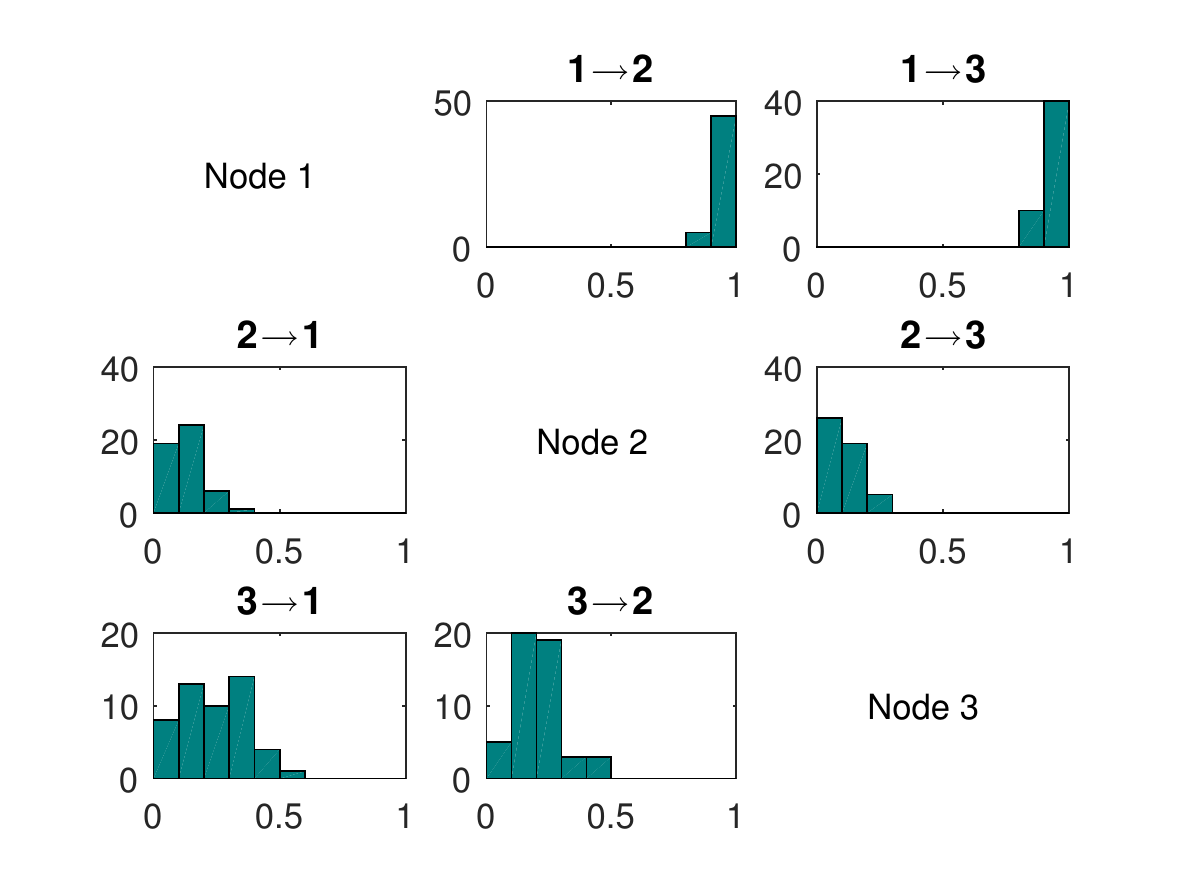}
		\caption{}
		\label{fig:hist_3fanout}
	\end{subfigure}%
	\begin{subfigure}{.5\textwidth}
		\centering
		\includegraphics[width=1\linewidth]{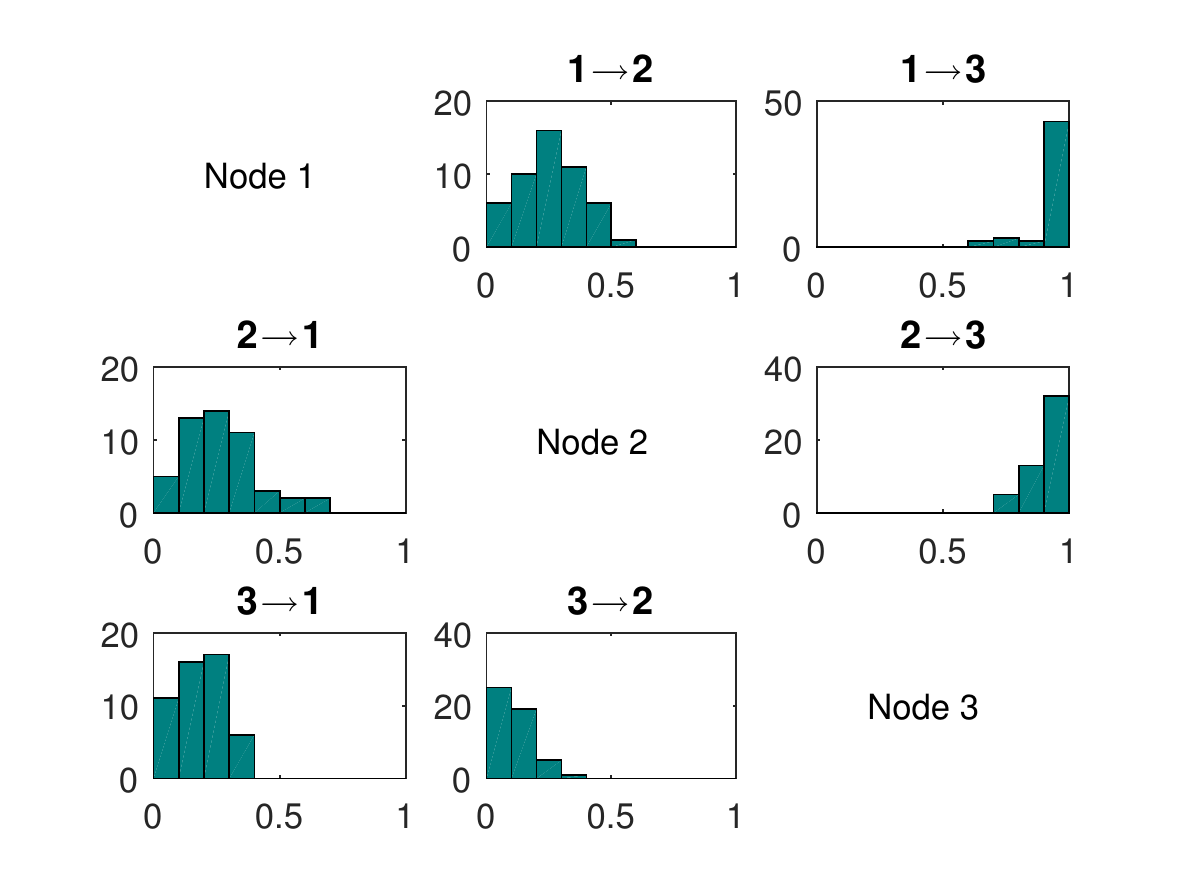}
		\caption{}
		\label{fig:hist_3fanin}
	\end{subfigure}
	\caption{Histogram of scores (normalized between 0 to 1) obtained by lsNGC for the (a) 3-fan out, (b) 3-fan in networks over 50 different sets of the simulation. (a) The influence of $x_1$ on $x_2$ and $x_3$ is captured quite well across the 50 simulations. (b) The influence of $x_1$ and $x_2$ on $x_3$ is captured quite well across the 50 simulations}
	\label{fig:hist_3}
\end{figure}

\begin{figure}[h]
	\centering
	\includegraphics[scale=0.7]{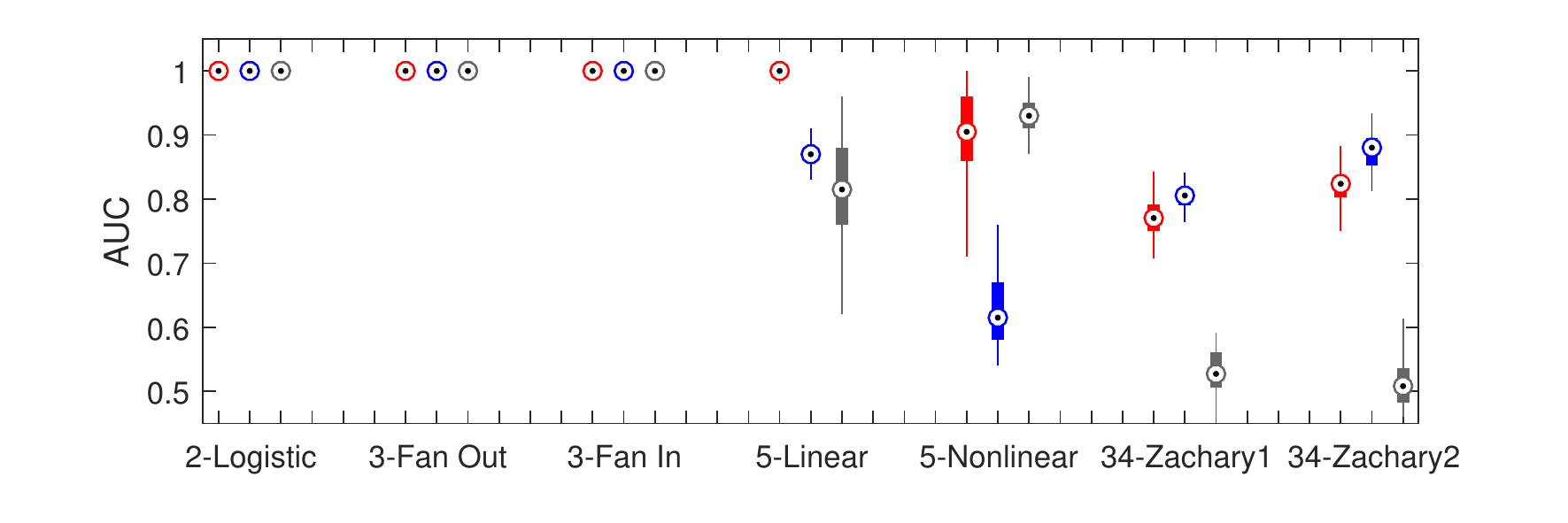}
	\caption{AUC results for the various networks comparing different methods, visualized as boxplots. The bottom end of the box represents the first quartile and the top of the box represents the third quartile. The circle with a dot in the box represents the distribution median. A general trend that is noticeable here is that the performance drops for all approaches as the number of nodes increases. Note that lsNGC performs competitively for all networks with LM slightly inferior for 5-linear and KGC significantly worse for all non-trivial networks.}
	\label{fig:AUC_res}
\end{figure}

\begin{figure}[h]
	\centering
	\includegraphics[scale=0.7]{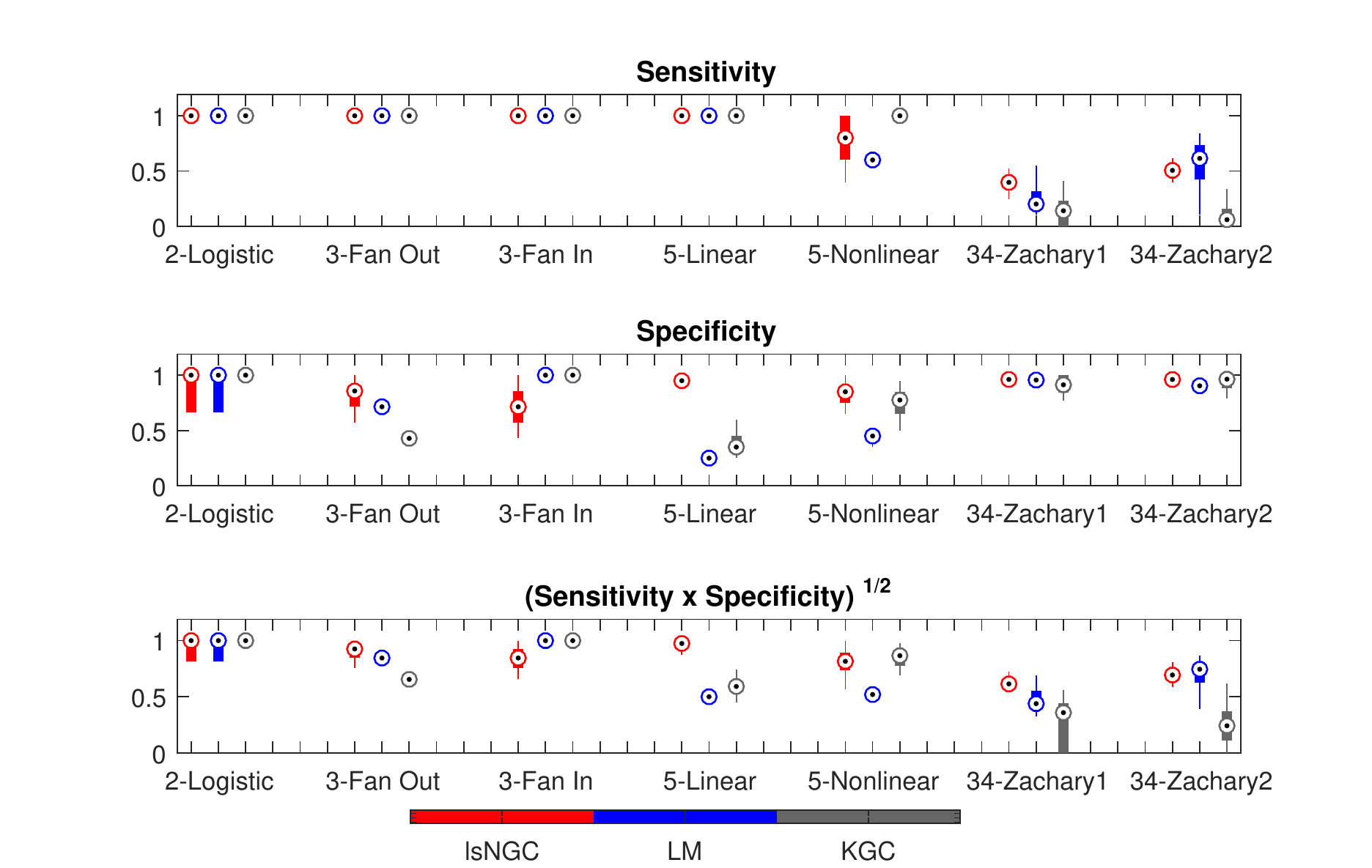}
	\caption{ Sensitivity and specificity results for the three approaches across all networks is plotted in this figure as boxplots. The bottom end of the box represents the first quartile and the top of the box represents the third quartile. The circle with a dot in the box represents the distribution median. Significant connections for LM was obtained by estimating LM measures between non-interacting pairs of surrogate time-series, generating a null distribution (see supplementary material). We also plot a measure combining the two measure to estimate overall performance. In general we observe that KGC performs most poorly.
	}
	\label{fig:sens_spec}
\end{figure}

\begin{figure}[h]
	\centering
	\includegraphics[scale=0.7]{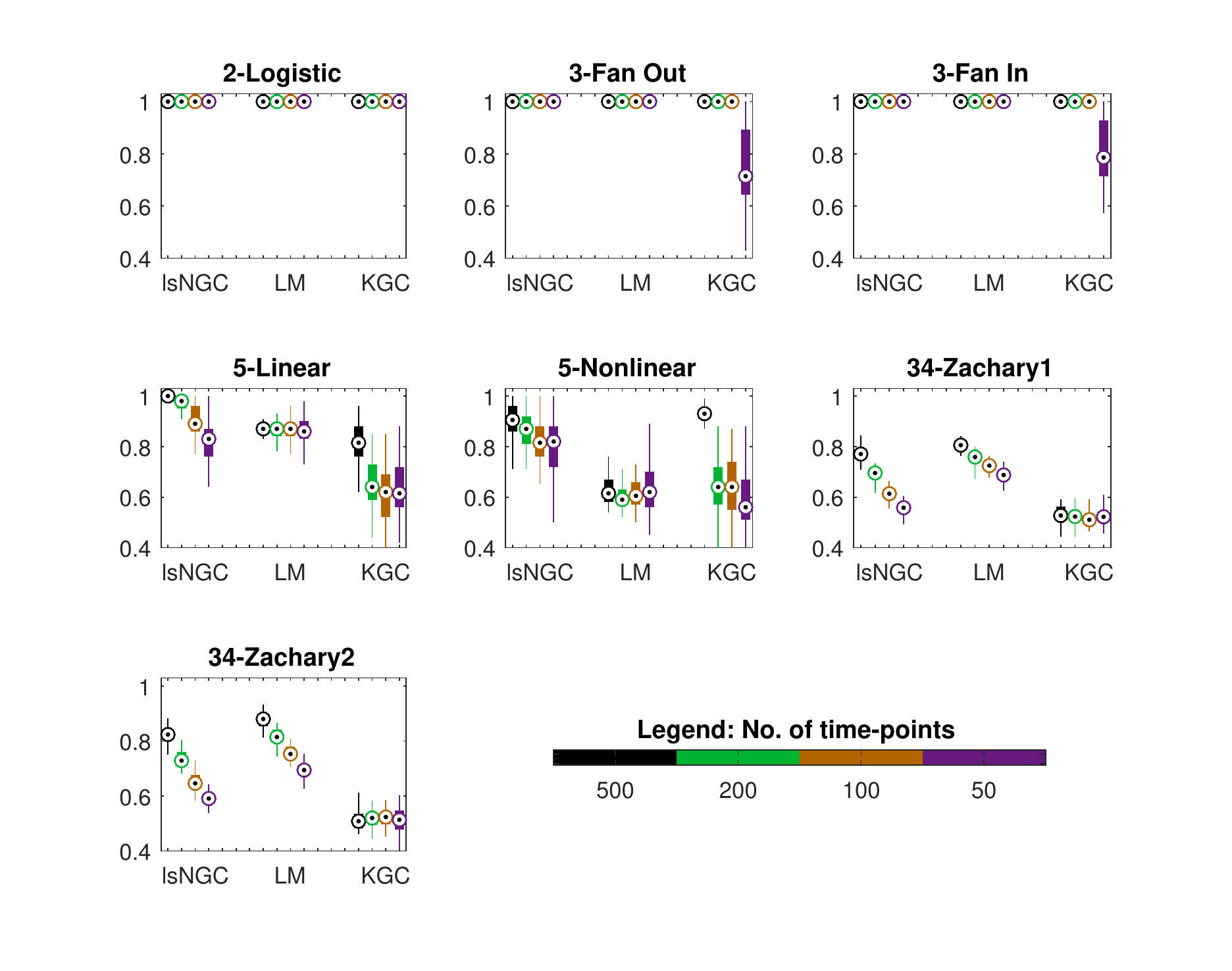}
	\caption{The effect of time-series length on performance of lsNGC, LM and KGC is studied here. We observe that for large networks, KGC requires a lot samples than LM or lsNGC to effectively recover the true network structure. LsNGC, on the other hand, performs well, with its performance reducing with fewer samples. LM is comparable to lsNGC; however, its does not perform too well for the 5-node network. The results are visualized as boxplots. The bottom end of the box represents the first quartile and the top of the box represents the third quartile. The circle with a dot in the box represents the distribution median.
	}
	\label{fig:time_lengths}
\end{figure}

In the fan-out case, nodes $x_2$ and $x_3$ are both driven by a common source, node $x_1$, hence the dynamics of the two driven nodes contain information from $x_1$. Thus, although $x_2$ and $x_3$ do not causally influence each other, they may be correlated. Such motifs can be challenging and an approach that is able to characterize these connections well is desirable. LsNGC is able to capture the connections well and is able to recover the fan-out network structure (Figure \ref{fig:hist_3fanout}). Figure \ref{fig:hist_3fanout}, clearly demonstrates that the scores assigned across all 50 simulations by lsNGC for $x_1 \to x_2$ and $x_1 \to x_3$ are much higher than any of the other connections. Additionally, no spurious connection is estimated between $x_2$ and $x_3$. 

The challenge faced when estimating connections with fan-in motifs is that since $x_3$ is influenced by $x_1$ and $x_2$, detected relationships are generally weak, since the dynamics of $x_3$ is affected by two time-series. From Figure \ref{fig:hist_3fanin}, we observe that the highest strengths of connection across all 50 simulations is rightly assigned to $x_1 \to x_3$ and $x_2 \to x_3$. However, we do observe lsNGC assigns relatively high strengths ($\sim$0.5) to $x_1 \to x_2$ and $x_2 \to x_1$ for a few of the 50 simulations. It would be interesting to dive into presence of such connections in more detail. We suspect this happens as a consequence of a multivariate model, since the v-structure \cite{koller2009probabilistic} ($x_1 \to x_3 \leftarrow x_2$) gets activated when $x_3$ is observed and information about $x_2 \ (x_1)$ is gleaned from $x_1 \ (x_2)$ if $x_3$ is observed. Nevertheless, as can be seen from Figure \ref{fig:hist_3}, the scores of the true connections are still higher than those of such spurious connections with only minimal overlap of the histograms. 

\paragraph{Five node nonlinear network:}

We also generate time-series data similar to that described in \cite{baccala2001partial} using the KGC toolbox (https://github.com/danielemarinazzo/KernelGrangerCausality). This toolbox contains both linear and non-linear implementations of interactions between time-series. Here, we present equations governing the non-linear 5-node network (5-nonlinear), while equations for the linear system (5-linear) are provided in the supplementary material.

\begin{equation}
\begin{gathered}
x_1(t) = x_1(t) + 0.95\sqrt{2}x_1(t-1)  - 0.9025x_1(t-2) \\
x_2(t) = x_2(t) + 0.5x_1^2(t-2) \\ 
x_3(t) = x_3(t) - 0.4x_1(t-3) \\
x_4(t) = x_4(t) - 0.5x_1^2(t-2) + 0.5\sqrt{2}x_4(t-1) + 0.25\sqrt{2}x_5(t-1) \\
x_5(t) = x_5(t) - 0.5\sqrt{2}x_4(t-1) + 0.5\sqrt{2}x_5(t-1) 		  
\end{gathered}
\end{equation}

Results estimating direction of connection are provided in the supplementary material (section 4).

\paragraph{34 node Zachary network:}
\label{zach}
Systems in nature involve of a number of interacting factors. Hence, it is important to evaluate systems with a considerably large number of interacting time-series. To test lsNGC on networks with a large number of nodes, we consider the Zachary dataset \cite{zachary1977information} consisting of 34 nodes. The nodal interactions is as follows and adopted from \cite{marinazzo2008kernel}:

\begin{equation}
\label{eq:zach}
x_{i}(t) = \Big(1 - \sum_{j=1}^{n}{c_{i,j}}\Big)(1-ax^2_{i}(t-1)) + \sum_{j=1}^{n}{c_{i,j}}(1-ax^2_{j}(t-1)) + s\tau_{i}(t); ~i = 1,~2, ~..., ~34
\end{equation}

Here, $a$ = 1.8, $s$ = 0.01, $c$ = 0.05, where $c_{i,j}$ represents the influence $j$ has on $i$, and $\tau$ is Gaussian noise with unit variance and zero mean. These quantities were adopted from \cite{marinazzo2008kernel}, where, the authors construct directed networks by assigning an edge, with equal probability of being in either direction. Apart from the directed connections, we randomly select 5 edges to be bidirectional. We construct 50 such networks and obtain 50-sets of time-series data from the corresponding network (34-Zachary2). One of the 50 networks used is shown in Figure \ref{fig:networks}. From the generated time-series data we estimate the underlying network structure of the 50 different networks. We also construct another 50 sets of time-series using the original undirected Zachary network with $c=0.025$ in equation (\ref{eq:zach}) (34-Zachary1).

\subsection *{Evaluating network recovery of simulations}
\label{evaluate_sims}
LsNGC derives measures of nonlinear connectivity scores represented as edges in a network. These are non-binary scores, from which we obtain a measure of the Area Under the receiver operating characteristic Curve (AUC). However, before deriving AUC measures, the connectivity matrix is log transformed to reduce the skew in the $f$-statistic measures. The Receiver Operating Characteristic (ROC) plots the true positive rate (TPR) versus the false positive rate (FPR), which shows the tradeoff between these quantities. Ideally, TPR should be 1 and FPR = 0 at any one threshold applied on the connectivity graphs, i.e. affinity matrix, for the AUC to equal 1. An AUC of 0.5 represents assignment of random connections, analogous to guessing the absence or presence of connections. Since the AUC quantifies both, the strength of connections and the direction of information flow, it is used to evaluate performance in estimating the true network structure. The AUC derives evaluation measures from the non-binarized connectivity matrix. However, it is also important to evaluate the statistical significance of connections i.e. edges in the network. The lsNGC measures of connectivity, expressed as $f$-statistic values, can be used to derive $p$-values for connections. Significant connections are obtained after multiple comparisons correction using False Discovery Rate (FDR) method at $p < 0.05$. From the thresholded affinity matrix, measures of $sensitivity$ and $specificity$ are derived.

Here, we present quantitative results on the recovered network structure for the various simulations. For every network investigated in this study, 50 different sets of time-series were simulated. Results are summarized as boxplots (example: Figure \ref{fig:AUC_res}, \ref{fig:sens_spec}). The circle with a dot inside the box represents the distribution median. The box spans the first quartile to the third quartile which is its interquartile range (IQR). The vertical extensions from the box, whiskers, have a maximum length of 1.5 times the IQR. The median of the AUC, sensitivity and specificity are represented as $\tilde{AUC}_{method}$, $\tilde{sens}_{method}$ and $\tilde{spec}_{method}$, respectively, where $method$ refers to the analysis method, i.e., lsNGC, LM or KGC. 

Results in Figure \ref{fig:AUC_res} were obtained for all network structures in Figure \ref{fig:networks} generated with 500 time-points. In this figure, red, blue and grey correspond to lsNGC, LM and KGC respectively. LsNGC, LM and KGC work very well for the smaller networks i.e. 2-logistic, 3-Fan out and 3-Fan-in networks. However, KGC's and LM's performance drops when using the linear system with 5 nodes, with a $\tilde{AUC}_{KGC} = 0.82$ and $\tilde{AUC}_{LM} = 0.87$, compared to $\tilde{AUC}_{lsNGC} = 1$. Additionally, LM performs poorly for the nonlinear 5 node network $\tilde{AUC}_{LM} = 0.62$, while lsNGC and KGC perform comparably. For both, directed and undirected 34-node Zachary networks, performance of lsNGC drops compared to its performance for smaller networks. It is, however, comparable to bivariate LM. The multivariate KGC performs poorly, with most recovered networks being random at medians of 0.52 and 0.51 for the networks. KGC performs poorly since it is cannot capture right connections for a relatively large network with just 500 time-points. In the original paper \cite{ma2014detecting}, the authors tested the Zachary network, but with 10,000 time-points; which is an unrealistic scenario for most practical applications.

Given, the F-statistic for the lsNGC measure, we obtain significant connections in the recovered network as described in section \ref{lsNGCexp}. Figure \ref{fig:sens_spec} plots the sensitivity, specificity and a measure combining them. Here, we observe that for small networks with 2-3 nodes, all approaches perform well with the exception of KGC for the fan-out network. For the 5-node networks, LM performs quite poorly, whereas overall, lsNGC does better than the two. Large networks, such as the Zachary network with 34 nodes, are generally difficult to recover since the total number of possible connections grows as a function of $N(N - 1)$. Here, we observe that KGC is the poorest of the methods tested. LM and lsNGC are comparable. However, it should be stressed that the calculation of significant connections for LM is not as straightforward as compared to lsNGC. Since a statistical measure cannot be derived directly from the score assigned by LM, surrogate time-series need to be created, from which LM measures are obtained, creating a null distribution. Details of this can be found in the supplementary material section 6.

Due to various constraints, such as cost, sensor limitations, manpower, time, etc., it is not always feasible to collect a small number of observations (time-points) for the factors under investigation. Thus, it is also essential to test the network recover-ability of lsNGC for fewer number of observations. To this end, the time-series length is varied from 500 to 50 time-points. Figure \ref{fig:time_lengths} compares AUC results across methods for decreasing number of time-points. Clearly, of the three approaches, KGC performs the poorest. For small networks, having 2-3 nodes, all methods perform comparably; however, a sudden drop in performance of KGC occurs when the time-series length reduces to 50 time-points for the 3 node networks. It is interesting to note that the sudden drop in KGC's performance is observed much earlier, at 200 time-points, for the 5-node network, while for the 34-node Zachary network, its performance oscillates across a median AUC of 0.5, indicating detection of random connections for 500 to 50 time-points.

LsNGC and LM perform equally well for the networks with 2-3 nodes across different time-series lengths. For the 5-node network, results indicate that LM reaches a bottleneck in performance, and is not able to improve as much as lsNGC can with increased time-series length. LsNGC has an AUC = 1 for the 5-node linear network for $T = 500$, whereas, the best performance with LM is a median AUC = 0.87. Nevertheless, it is important to note that network structure recovered with LM does not degrade with decreasing time-series lengths. For the 5-node nonlinear network, we see that the best performance with LM is $\tilde{AUC}_{LM} = 0.62$, whereas the worst for lsNGC is $\tilde{AUC}_{lsNGC} = 0.82$. Additionally, we investigate both undirected (34-Zachary1) and directed networks (34-Zachary2). The drop in performance of lsNGC is a lot steeper than that of LM with decreasing time-series lengths. However, its performance is comparable to LM at $T > 200$. 

\subsection{Functional Magnetic Resonance Imaging data}
LsNGC showed promising results on the simulations. Nevertheless, its performance on real data can give more insight into its usability for various applications involving network graph estimation from signals. In this work, we analyze its performance on functional Magnetic Resonance Imaging (fMRI) data. It has been demonstrated that individuals presenting with symptoms of HAND have quantifiable differences in connectivity \cite{thomas2013pathways,abidin2018alteration} from controls. We hypothesize that if lsNGC can capture brain connectivity from fMRI data for the subjects well, differences in connectivity between subjects with HAND and controls should be observed. Hence, we tested how well a classifier was able to discriminate the two subject groups. The classifier is able to learn relevant differences from the two groups using connectivity derived with lsNGC (AUC =0.88 and accuracy = 0.77) suggesting that lsNGC was able to characterize the network structure well. More details on the data and analysis approach can be found in appendix C.

\section{Discussion}

In this paper we introduce a novel approach called large-scale nonlinear Granger causality (lsNGC). This approach is a nonlinear, multivariate variant to Granger causality that can estimate connections in systems with a large number of time-series. We demonstrated its applicability on various simulations and on real data. In this section, we discuss the performance of lsNGC in instances where it effectively recovered network structure, as well as in simulations where it did not. All investigated methods, i.e. lsNGC, LM andnKGC have in common that for a given number of observations (time-points) their performance drops with increasing number of nodes. However, performance may be improved by increasing the number of observations. We observe that lsNGC outperforms the two other nonlinear approaches in most cases. Further, KGC is a lot more susceptible to poor performance with increased number of nodes (time-series, Figure \ref{fig:time_lengths}) compared to lsNGC.. When increasing the number of nodes in the network, the number of time-points has to be significantly increased to produce meaningful results with KGC. Although, lsNGC's performance gradually drops with decreasing number of time-points, for all practical lengths of time-series, it outperforms KGC, making lsNGC more reliable for larger systems with fewer time-points.

Comparing lsNGC with LM, it is seen that the network structure recovered with LM does not degrade as rapidly with decreasing time-series lengths. This can be attributed to the lesser complexity of LM, hence fewer parameters to be estimated compared to lsNGC. As such, given very short time-series LM may be able to outperform lsNGC. LM, having fewer parameters, is a less complex model but with high bias. Such high bias comes at a the price of significant performance drop of LM, which is clearly demonstrated in the 5-node nonlinear network as a good example. Here, we see that LM performs best with  $\tilde{AUC}_{LM} = 0.62$, whereas the worst result for lsNGC is $\tilde{AUC}_{lsNGC} = 0.84$. To put it simply, LM is a low variance, high bias model, whereas lsNGC is a higher variance, lower bias model. This becomes more evident when analyzing the network with 34 nodes. The drop in performance of lsNGC is a lot steeper than that of LM with decreasing time-series lengths. However, its performance is comparable to LM at $T > 200$. Additionally, the formulation of lsNGC can be directly used to estimate significant connections, unlike LM, where a null distribution needs to be created from a set of surrogate time-series (Supplementary material section 5). This increases the computational cost to a large extent. The flexibility of estimating significant connections with lsNGC is a significant advantage over traditional approaches for detecting causality. 

The success of lsNGC on simulated data motivated us to test its performance on real data. To this end, we evaluated the connectivity matrices derived using lsNGC on real fMRI data from healthy controls and subjects with HIV associated neurocognitive disorder (HAND). The connectivity measures used as features in a classifier were highly discriminative. This demonstrates that lsNGC is able to capture relevant information regarding interaction between different regions in the human brain.

In summary, lsNGC is quite robust in recovering network structure across different network architectures. Similar to all investigated methods, network size does affect its performance; however, it significantly outperforms conventionally used methods such as KGC for practical time-series lengths (Figure \ref{fig:time_lengths}). In comparison, as LM is a bivariate method, it does not consider confounding time-series in its models, and falls short at capturing indirect connections. LsNGC has the benefit of being a nonlinear, multivariate  method whose formulation provides control over the number of parameters to be estimated, as such, lsNGC can effectively estimate network structure even in high-dimensional systems with short time-series.  

\section{Conclusion}

In this work, we introduce an approach for estimating underlying directed network structure from time-series data. We propose a multivariate, nonlinear method called large-scale nonlinear Granger causality (lsNGC) for detecting causal interactions between time-series. Most approaches proposed in literature for performing multivariate nonlinear causality analysis are limited by the number of samples of the time-series. Additionally, few multivariate approaches can handle large number of nodes in the network due to computational limitations, making them impractical for large networks. The practicality and limitations of methods can be investigated through experimentation and analysis with different types of networks. In this study, we demonstrated the advantage of lsNGC over traditional multivariate and bivariate approaches to recovering directed network graphs. The high AUC, good sensitivity and specificity results for realistic lengths of time-series data demonstrates its potential to be applied to real world data. Additionally, although the bivariate LM performs comparable to lsNGC in some instances, obtaining the true, unweighted network structure using LM is difficult since it requires the calculation of surrogate time-series, which is computationally very expensive, especially for large networks. Finally, we have demonstrated the applicability of lsNGC in inferring brain network graphs from brain activity data obtained using functional magnetic resonance imaging (fMRI). Besides clinical applications for diagnosis of neurological disorders, such an approach may be able to reveal useful insights about directed interactions in the brain.

\section*{APPENDIX}

\section*{A. Granger causality analysis}

The principle of Granger causality (GC) is based on the concept of precedence and predictability, where the improvement in prediction quality of a time-series in the presence of the past of another time-series is evaluated and quantified, revealing the directed influence between the two series \cite{Granger1969} investigated. As lsNGC is an extension of traditional multivariate GC (mvGC), the basic concepts of mvGC are briefly described here.

Consider a system with $N$ time-series, each with $T$ temporal samples. Let the time-series ensemble \- $\bcx \in \mathbb{R}^{N \times T}$ be 
${\bcx = (\bx_1, \bx_2, ..., \bx_N)^T}$, where $\bx_n \in \mathbb{R}^{T}$, $ n \in \{1, 2, ..., N\}$, $\bx_n = (x_n(1), x_n(2), ..., x_n(T))$. The time-series ensemble $\bcx$ can also be represented as $\bcx = (\bx(1), \bx(2), ..., \bx(T))$, where $\bx(t) \in \mathbb{R}^{N \times 1}$, \-$t \in \{1, 2, ..., T\}, \bx(t) = (x_1(t), x_2(t), x_3(t), ..., x_N (t))^T$.

We use multivariate vector auto-regression which is the most common prediction scheme used in GC analysis \cite{Ryali2011, Granger1980}. 

\begin{equation}
\label{eq:GC_orit}
\bx(t) = \sum_{j=1}^{d}{\bca_j\bx(t-j)+\be(t)}       
\end{equation}
Here, the matrices $\bca_j$ are the model parameters obtained by minimizing the mean squared errors in the estimate of $\bcx$, where $\bca_j$ is an $N \times N$ matrix with $j \in {1, 2, ..., d}$, and $d$ is the lag order. We define $\hat{\bcx}$ as the predicted system (1) without the error term, and $\bce$ as the set of residuals defining the difference between the actual and predicted values of $\bcx$.

\begin{equation}
\label{eq:GC_unrest}
\hat{\bx}(t) = \sum_{j=1}^{d}{\bca_j\bx(t-j)}   
\end{equation}
\begin{equation}
\bce = \bcx - \hat{\bcx}   
\end{equation}

Granger causality (GC) analysis establishes a causal influence score from time-series $\bx_s$ to $\bx_r$ on the premise that, if the predictability of time-series $\bx_r$ improves in the presence of the past of time-series $\bx_s$, then $\bx_s$ Granger causes $\bx_r$. GC analysis estimates causal relationships in a multivariate sense by considering the full ensemble of time-series including confounding time-series that neither represent $\bx_s$ nor $\bx_r$. We obtain the influence of $\bx_s$  on $\bx_r$ by quantifying the reduction in prediction quality of $\bx_r$  in the absence of time-series $\bx_s$. Equations (2) and (3) obtain the prediction of all series when the full ensemble of time-series is used. Next, we obtain the prediction error $\bce_{\bcx\backslash\bx_s}$, when series $\bx_s$ is removed from the ensemble $\bcx$.

If the prediction quality of $\bx_r$ is higher when $\bx_s$ is used (2) rather that when it is not used for predicting time-series $\bx_r$, then it is said that $\bx_s$  Granger causes $\bx_r$. The $f$-statistic can be obtained by recognizing that the two models can be characterized as the unrestricted model and the restricted model, where the unrestricted model is eq. (\ref{eq:GC_unrest}), and the restricted model is constructed in the absence of time-series $\bx_s$. Residual sum of squares ($RSS$) of the restricted model, $RSS_{R}$, and residual sum of squares of the unrestricted model $RSS_{U}$ are obtained. $n$ is the number of observations in the regression, $q_{U}$ and $q_{R}$ are the number of parameters to be estimated for the unrestricted and restricted model, respectively. For traditional multivariate GC, $q_{R} = d\times (N-1)$  and $q_{U} = d\times N$, respectively. 



A measure of GC can be obtained using the $f$-statistic, given by:
\begin{equation}
\label{eq:fGCtest}
FGC_{\bx_s\to\bx_r}  = \frac{(RSS_{R} - RSS_{U})/(q_{U} - q_{R})}{(RSS_{U})/(n - q_{U} - 1)}
\end{equation}

$FGC_{\bx_s\to\bx_r}$ quantifies the influence of $\bx_s$ on $\bx_r$, by testing the equality of variances of errors in prediction of the $\bx_r$ by both the models. If the variance of the error in predicting $\bx_r$ is lower when $\bx_s$  is used, then $\bx_s$ Granger causes $\bx_r$.  Significant interactions can be obtained using the $f$-statistic.

\section*{B.Implementation specifics for large-scale Nonlinear Granger causality (lsNGC)}

\subsection*{Implementation specifications}
\label{Implement}

Nonlinear Granger causality analysis has been investigated and theoretical work laying the foundation of mathematical formulation to perform such an analysis has been studied \cite{ liao2009kernel,marinazzo2011nonlinear, li2010nonlinear}. However, while sound theoretical concepts are prerequisite, practical implementation, especially for large systems having many nodes (time-series) is not always straightforward. In this section, we briefly describe implementation specifications that increase scalability and reduce computation time significantly.

Let us say that we are interested to learn if $\bx_s$ influences $\bx_r$. We first construct the phase space representation of $\bx_s$ with embedding dimension $d$, as $\bm{W}_s$. The state at time $t$ is $\bm{w}_s(t) = [x_s(t-(d-1)), ..., x_s(t-1), x_s(t)]$, and $t \in {d, ..., T-1}$. 

To perform a multivariate analysis, a phase space reconstruction is constructed, where prediction is performed using all the time-series apart from $\bx_s$ whose influence is to be quantified. From the time-series ensemble $\bcx\backslash\bx_s$ we construct the phase space reconstruction $\bm{Z}_s$. The state of this multivariate system at a given time-point is 

\[
\bm{z}_s(t) =\left\{
\begin{array}{ll}
x_1(t-(d-1)), ..., x_1(t-1), x_1(t), ...\\
x_2(t-(d-1)), ..., x_2(t-1), x_2(t), ...\\	
\tab .\\
\tab .\\
x_{N-1}(t-(d-1)), ..., x_{N-1}(t-1), x_{N-1}(t)\\
\end{array}
\right.
\]
It should be noted that $\bm{Z}_s$ does not contain any terms from $\bx_s$. Let $\mathbf{f}$ and $\mathbf{g}$ represent two nonlinear functions. The two estimates of $\bcx$ are given by:

\begin{equation}
\label{eq:unrestrct}
\hat{\bcx}=\mathbf{A}_{1}\mathbf{f}(\bm{Z}_s) + \mathbf{A}_{2}\mathbf{g}(\bm{W}_s)
\end{equation}
\begin{equation}
\label{eq:restrct}
\tilde{\bcx}\backslash\bx_s=\mathbf{B}_{1}\mathbf{f}(\bm{Z}_s) 
\end{equation}

In the above equations, $\mathbf{A}_{1}$, $\mathbf{A}_{2}$ and $\mathbf{B}_{1}$ are the weights or model parameters, obtained by minimizing the mean squared errors in the estimate of all time-series in $\bcx$. The time-series $\hat{\mathbf{x}}_{r, s}$ from $\hat{\bcx}$ and ${\tilde{\mathbf{x}}}_{r,s}$ from $\tilde{\bcx}\backslash\bx_s$ are the estimates obtained for $\bx_r$. The subscript $(r, s)$ denotes that these are estimates of $\bx_r$ obtained to investigate the influence of $\bx_s$ on $\bx_r$. The two constructed models, given in eq. (\ref{eq:unrestrct}) and (\ref{eq:restrct}), are the unrestricted and restricted model, respectively. The RSS of the two models gives us an $f$-statistic measure. In this study, we use the generalized radial basis function (GRBF) neural network as nonlinear transformations $\mathbf{f}$ and $\mathbf{g}$.

\begin{algorithm}[H]
	\label{algo_lsNGC}
	\caption{Large-scale nonlinear Granger causality algorithm}\label{lsNGC_algo}
	\begin{enumerate}
		\item Normalize the time-series to have zero mean and unit standard deviation focusing solely on system dynamics.
		\item Using the order $d$, obtain phase space reconstructions of $\bm{Z}$. Here, $\bm{Z}$ is obtained using all the time-series in the system, $\bm{Z} \in \mathbb{R}^{Nd \times (T-d)}$. 
		\item From $\bm{Z}$, obtain $c_f$ number of cluster centers with $k$-means clustering. Cluster centers  $\bm{U}$ can be thought of as parameters of the hidden layer of a Generalized Radial Basis Function (GRBF) network having dimensions ${Nd \times c_f }$. 
		\item Set the width of the kernel as the mean distance between cluster centers.
		\item Iterate through all $N$ time-series from 1 to $n$ where $n \in N$, selecting one time-series ($\bx_s$) whose influence on other time-series is to be investigated.
		
		\begin {enumerate}
		\item Obtain phase space reconstructions $\bm{Z}_s \in \mathbb{R}^{(N-1)d \times (T-d)}$. States in this phase space do not contain information about $\bx_s$.
		\item Obtain $\bm{W}_s \in \mathbb{R}^{d \times (T-d)}$, the phase space reconstructions of $\bx_s$, having embedding dimension $d$.
		\item Obtain $c_g$ number of cluster centers in the phase space $\bm{W}_s$ with $k$-means clustering. Set the width of the kernel as the mean distance between cluster centers. These parameters, $\bm{V}$, have dimensions ${d \times c_g}$.
		\item Calculate activations for each of the $c_g$ neurons using equation (\ref{grbf_activations}), given by $\mathbf{g}(\bm{W}_s)$, from each of the states in $\bm{W}_s$.
		\item From $\bm{U}$, eliminate those dimensions corresponding to $\bx_s$. For example, if $\bx_s$ is the first time-series in the system, eliminate the first $d$ dimensions of the cluster centers $\bm{U}$. In general, if it is the $n$-th time-series, eliminate $d$ indices starting from and including $nd-(d-1)$. This results in cluster centers $\bm{U}_s$.
		\item Calculate activations for each of the $c_f$ neurons using equation (\ref{grbf_activations}), given by $\mathbf{f}(\bm{Z}_s)$, from each of the states in $\bm{Z}_s$.	
		\item \textbf{Predictions in the presence of} $\bx_s$: Obtain $\hat{\bcx}$	
		\item \textbf{Predictions in the absence of} $\bx_s$: Obtain $\tilde{\bcx}\backslash\bx_s$
		\item Calculate the influence of $\bx_s$ on every time-series in $\bcx$ using the $f$-statistic.
		\end {enumerate}	
	\end{enumerate}
\end{algorithm}


\subsection*{Nonlinear transformation using Generalized Radial Basis Function}

In this work we adopt the Generalized Radial Basis Function (GRBF) as the nonlinear transformation $\mathbf{f}$ and $\mathbf{g}$. Cluster centers ${\bm{V}^\text{T} \in \mathbb{R}^{c_g \times d}}$ are calculated for the state space $\bm{W}_s$, where $c_g$ is the number of clusters obtained with $k$-means clustering. Activation function $\mathbf{g}$ in (5) is calculated as follows:

\begin{equation}
\label{grbf_activations}
g_i(\bm{w}_s(t))= \frac{e^{-{||\bm{w}_s(t) - \bm{v}(i)||}^2/  {\sigma}^2}}{{\sum}_{j=1}^{c_g}{e^{-{||\bm{w}_s(t) - \bm{v}(i)||}^2/  {\sigma}^2}}}
\end{equation}

where, $i \in$ \{1, 2 ... $c_g$\}

Analogously, cluster centers ${\bm{U}_s^\text{T} \in \mathbb{R}^{c_f \times (N-1)d}}$ are calculated for the state space $\bm{Z}_s$, where $c_f$ is the number of clusters obtained with $k$-means clustering. Activation function $\mathbf{f}$ in (5) and (6) is calculated as follows:

\begin{equation}
\label{grbf_f_activations}
f_i(\bm{z}_s(t))= \frac{e^{-{||\bm{z}_s(t) - \bm{u}_s(i)||}^2/  {\sigma}^2}}{{\sum}_{j=1}^{c_f}{e^{-{||\bm{z}_s(t) - \bm{u}_s(i)||}^2/  {\sigma}^2}}}
\end{equation}

where, $i \in$ \{1, 2 ... $c_f$\}

The embedding dimension $d$ is chosen using Cao's method described in \cite{cao1997practical}. In this study, $c_f$ = 25 and $c_g$ = 5 is chosen empirically from preliminary analysis. 

\section*{C. Data and Methods}
\subsection*{5-node linear network}

The linear implementation of interactions between the 5-node network time-series (whose network structure is the same as in example 3 of reference \cite{baccala2001partial}) is provided here:
\begin{equation}
\begin{gathered}
x_1(t) = x_1(t) + 0.95\sqrt{2}x_1(t-1)  - 0.9025x_1(t-2) \\
x_2(t) = x_2(t) + 0.5x_1(t-2) \\ 
x_3(t) = x_3(t) - 0.4x_1(t-3) \\
x_4(t) = x_4(t) - 0.5x_1(t-2) + 0.5\sqrt{2}x_4(t-1) + 0.25\sqrt{2}x_5(t-1) \\
x_5(t) = x_5(t) - 0.5\sqrt{2}x_4(t-1) + 0.5\sqrt{2}x_5(t-1) 		  
\end{gathered}
\end{equation}

\subsection*{5-node time-series results}

\begin{figure}[h]
	\centering
	\includegraphics[scale=0.6]{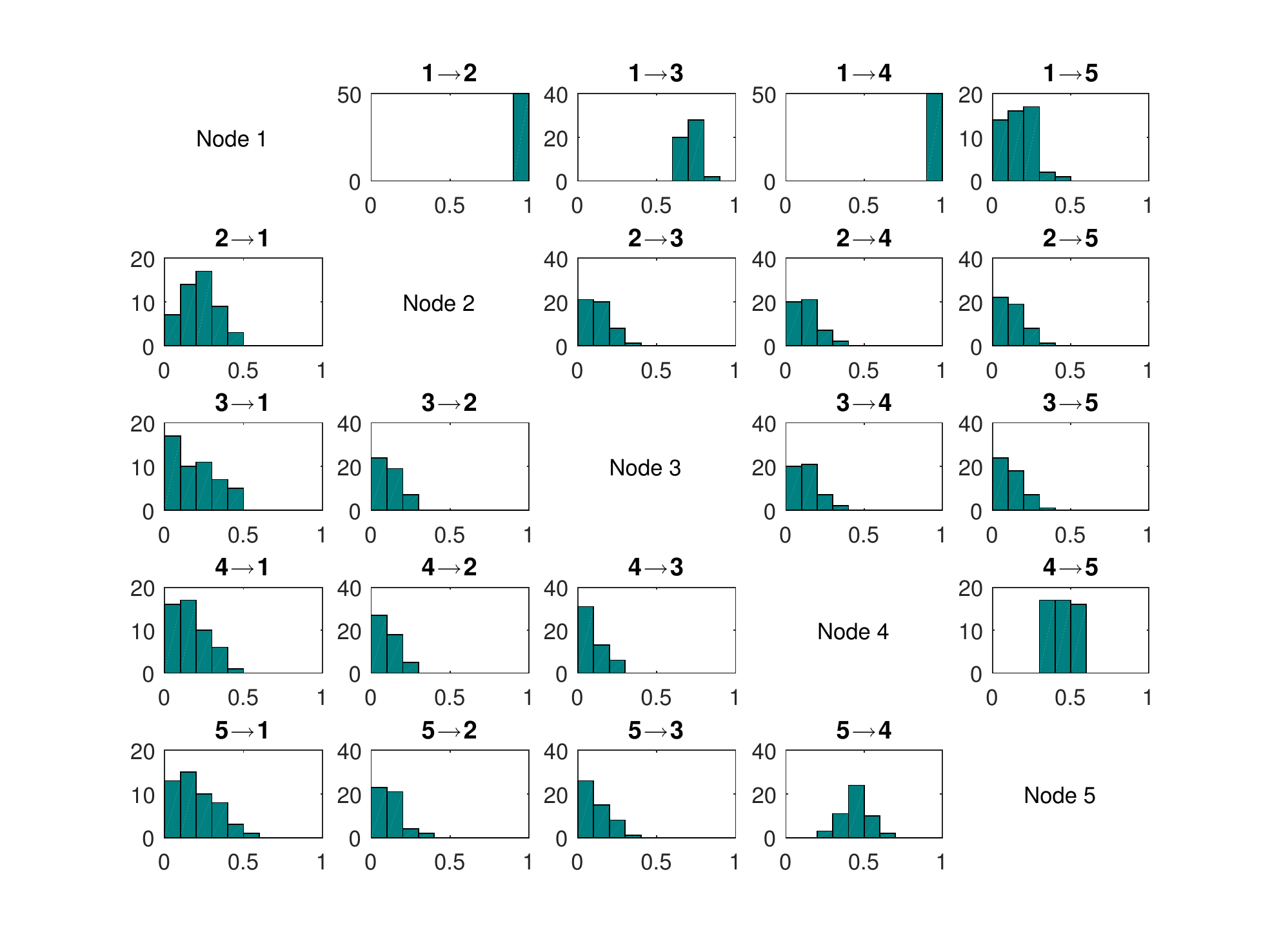}
	\caption{Histogram of scores (normalized between 0 to 1) obtained by lsNGC for the 5-node, linear (5-linear) network over 50 different sets of the simulation.}
	\label{fig:hist_5lin}
\end{figure}

The true connections of the 5 node network can be summarized as follows, $x_1 \to x_2,~ x_1 \to x_3,~ x_1 \to x_4,~ x_4 \to x_5$ and $x_5 \to x_4$, for the linear and non-linear case. LsNGC in the linear case clearly assigns high scores to the right connections, Figure \ref{fig:hist_5lin}. LsNGC correctly assigns a high score to $x_1 \to x_2,~ x_1 \to x_3,~ x_1 \to x_4, ~x_4 \to x_5$ and $x_5 \to x_4$ and assigns low scores to the rest of the connections which rightly correspond to non-existent connections. The high median AUC, specificity and sensitivity of 1, as shown in Figures 4 and 5 of the main paper, demonstrate that the correct network graph was recovered in most of the cases. LsNGC does not perform as well in the nonlinear case as compared to the linear one. Here, connections from $x_1 \to x_3$ and $x_4 \to x_5$ are detected well. However, the other connections were weaker. Quantitatively, this weak separation between the scores estimated by lsNGC for absence and presence of connections corresponds to the lower AUC, specificity and sensitivity values, as shown in Figures 4 and 5 of the main paper.

\newpage
\begin{figure}[H]
	\centering
	\includegraphics[scale=0.6]{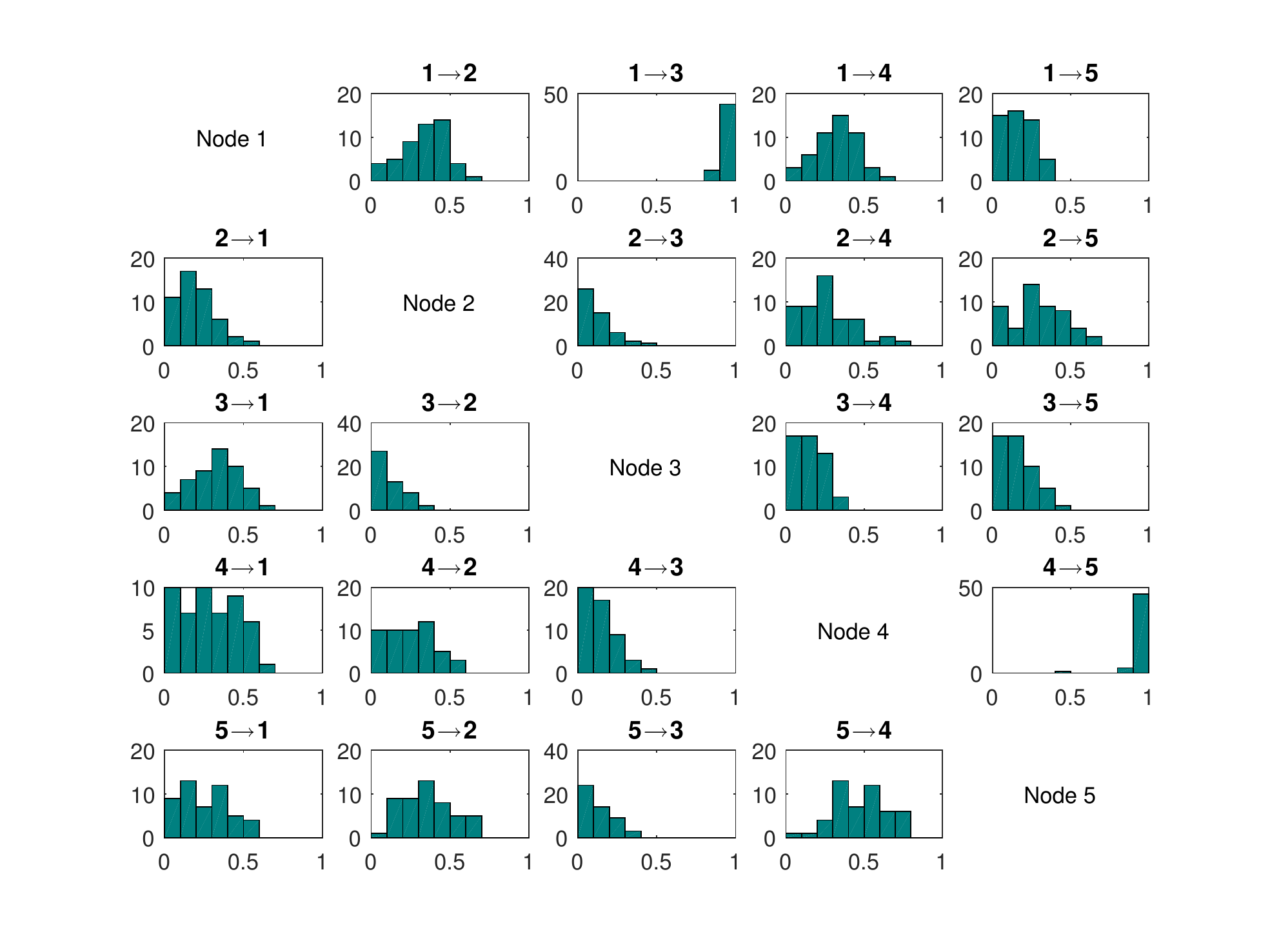}
	\caption{Histogram of scores (normalized between 0 to 1) obtained by lsNGC for the 5-node, nonlinear (5-nonlinear) network over 50 different sets of the simulation.}
	\label{fig:hist_5nonlin}
\end{figure}

\subsection*{Resting-state fMRI data:}
\label{empirical_data}

Functional MRI scans were obtained from human subjects at the Rochester Center for Brain Imaging (Rochester, NY, USA) using a 3T, Siemens Magneton TrioTim scanner. The study protocol included: (i) High-resolution structural imaging using T1-weighted magnetization-prepared rapid gradient echo sequence (MPRAGE, TE = 3.44 ms, TR= 2530 ms, isotropic voxel size 1 mm, flip angle = 7$^{\circ}$). (ii) Resting-state fMRI scans using a gradient spin echo sequence (TE = 23 milliseconds, TR = 1650 milliseconds, 96 $\times$ 96 acquisition matrix, flip angle of 84$^{\circ}$). The acquisition lasted 6 minutes and 54 seconds, and 250 temporal scan volumes were obtained. A total of 25 slices, each 5 mm thick, were acquired for each volume. During acquisition, the subject was asked to lie down still with eyes closed. The data were acquired as part of a NIH sponsored study (R01-DA-034977). Prior to computation of connectivity measures, the fMRI data used in this study was preprocessed using standard methodology. For each dataset, the first ten (of 250) volumes of functional magnetic resonance images were eliminated to analyze only those that reached steady-state imaging. Next, motion correction, brain extraction and correction for slice timing acquisition were performed. Additional nuisance regression was carried out to remove variations due to head motion and physiological processes. Each dataset was finally registered to the 2 mm MNI \cite{mazziotta2001probabilistic} standard space using a 12-parameter affine transformation \cite{jenkinson2001global}. All preprocessing steps were carried out using the C-PAC software \cite{sikka2014towards} and its corresponding dependencies in FMRIB Software Library (FSL) \cite{Smith2004}. Finally, the time-series were normalized to zero mean and unit standard deviation to focus on signal dynamics rather than amplitude \cite{Wismuller2002}. Based on the commonly used Automated Anatomic Labeling (AAL) template \cite{tzourio2002automated}, the registered MRI volumes were divided into 90 regions, excluding the brain stem and cerebellar regions, 45 in each hemisphere. A representative time-series for each region was computed by averaging the time-series of all voxels within it. 

Subjects in this study were recruited as part of a NIH funded study (R01-DA-034977) at the University of Rochester Medical Center. In total, 15 healthy controls (mean age 42 $\pm$ 10 years) and 14 HIV positive subjects with symptoms of HIV associated neurocognitive disorder (HAND, mean age 45 $\pm$ 16 years) were recruited as part of this study. A standard battery of neuropsychological (NP) tests was used to assess cognitive abilities of subjects, covering six cognitive domains: executive function, speed of information processing, attention, memory, learning, and motor function. These scores were converted to age and education adjusted z-scores. An overall z-score combining the scores from individual domains was generated and used to assess HAND \cite{antinori2007updated}. All participants provided written informed consent prior to participation as per protocol approved by the institutional IRB.


\begin{figure}[h]
	\centering
	\includegraphics[scale=1.0]{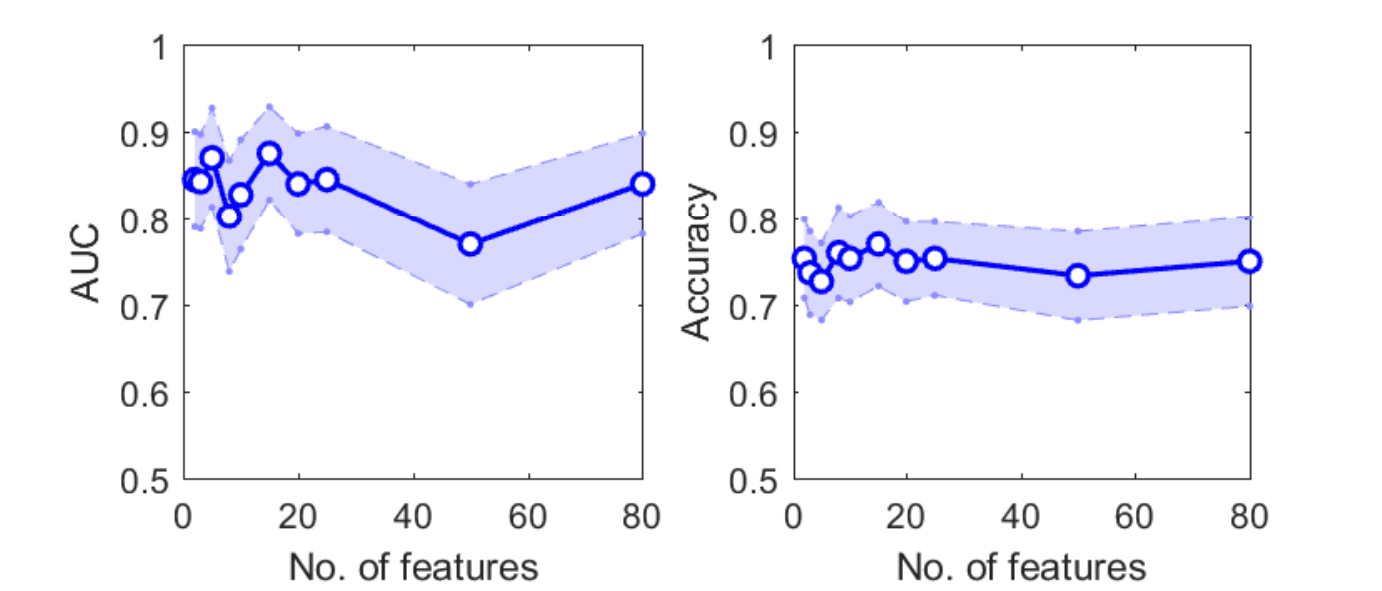}
	\caption{Plot of AUC and accuracy results for different number of retained features. Shaded regions above and below each solid line, corresponding to the mean AUC/accuracy, represent the 95\% confidence interval of the AUC/accuracy value. The general trend observed here is that lsNGC is able to discriminate between the two subject groups well.}
	\label{fig:AUC_accuracy}
\end{figure}

\paragraph{Application to fMRI data:}
\label{mvpa_proc}

It has been demonstrated that individuals presenting with symptoms of HAND have quantifiable differences in connectivity \cite{thomas2013pathways,abidin2018alteration} from controls. We hypothesize that if lsNGC can capture brain connectivity from fMRI data for the subjects well, differences in connectivity between subjects with HAND and controls should be observed. Hence, we tested how well a classifier was able to discriminate the two subject groups. First, matrices of interaction using lsNGC, connectivity matrices, were obtained from the fMRI time-series of every subject in this study. LsNGC produced connectivity profiles conveying different information about interactions amongst regional time-series for every subject. These connectivity matrices were vectorized such that data from each subject was represented as an $F$ dimensional vector of interactions, which can be viewed as features to train a classifier. Since, the number of features were large ($ \sim 8000$ for $N=90$, where $N$ is the number of regions defined by the AAL template) compared to the number of subjects in our study, before vectorizing the connectivity matrices, we symmetrized them such that $F=N(N-1)/2$, reducing the number of features to 4005 for each subject.

To further reduce the number of redundant and/or noisy features, we performed feature selection, using Kendall's $\tau$ correlation coefficient \cite{kendall1955rank} . Feature selection aims at estimating interactions that best discriminate subjects presenting with HAND symptoms and controls and reduces the computational complexity of a classifier. Feature selection was performed independently for every set of training/test set split. Feature ranking estimated by Kendall's coefficient feature selection approach was used to select $s$ features where $s \in \{2, 3, 5, 8, 10, 15,20, 25, 50, 80\}$.

The ranked features were classified with the AdaBoost \cite{freund1997decision} classifier. It is an ensemble classifier that uses an ensemble of weak classifiers, such as decision stump classifier, to produce a strong classifier.  The WEKA \cite{hall2009weka} implementation of AdaBoost was imported into MATLAB 2016 and used in our analysis. 

\label{pref}
We ensured a strict split between training and testing data, with $90\%/10\%$ train/test separation within an iterative cross-validation scheme with 100 different data splits. The training data was used for both feature selection and training the classifier. Classification accuracy and the AUC were adopted to evaluate performance. An AUC of 1 indicates perfect classification while an AUC of 0.5 indicates random classification. The classifier for discriminating subjects with HAND and healthy controls achieved best mean AUC = 0.88 and accuracy = 0.77 with 15 features (Figure \ref{fig:AUC_accuracy}) which suggests that lsNGC was able to characterize the network structure well, hence the classifier was able to learn discriminate characteristics well. 

\section*{D. Comparative methods for evaluating lsNGC}

In this section we briefly summarize the two comparison methods used to evaluate the performance of lsNGC.

\subsection*{Causality analysis with local models}

An example for the use of local models (LM) to estimate casual relationships between time-series in a system is described in \cite{sugihara2012detecting}. First, the attractor manifold of a time-series is constructed from time-delayed versions of itself and embedded into a state-space reconstruction (SSR). Such a space of embedded points transforms the observed time-series into a manifold \cite{takens1981detecting, deyle2011generalized} representing the evolution of states, i.e., its dynamics. The basic idea here is to build local models (using nearest neighbors) of the state space dynamics for every time-series, which cross-predicts the trajectory of an influencing time-series. Where, cross-prediction implies predicting say $x_1$ from the states of $x_2$. If $x_2$ cross-predicts $x_1$ well, then $x_1$ influences $x_2$.

\subsection*{Kernel Granger causality (KGC)}

Kernel Granger causality (KGC) was proposed in 2008 \cite{marinazzo2008kernel}. It calculates Granger causality in the feature space of kernel functions. In this work, we use radial basis function kernels for KGC. \cite{baccala2001partial} We used the publicly available KGC toolbox (https://github.com/danielemarinazzo/KernelGrangerCausality) to perform this analysis. 

\section*{E. Estimating significance of connections from networks obtained}

Measures of significance using lsNGC can be obtained from the $f$-statistic (equation (3)) and details regarding obtaining significance of connections with KGC can be found in \cite{marinazzo2008kernel}. Unlike lsNGC and KGC, LM measures cannot be used directly to glean significance measures. Significance is calculated by establishing a null distribution, which was obtained by estimating LM measures between non-interacting pairs of surrogate time-series. The surrogates were generated by using the Iterative Amplitude Adjusted Fourier Transform \cite{schreiber1996improved} algorithm generated with the Chaotic Systems Toolbox \cite{CST2}. 
Sensitivity and specificity were calculated on the connectivity matrix thresholded at $p$ < 0.05, followed by multiple comparisons correction using False Discovery Rate, where the null hypothesis is generated by estimating connections between non-interacting surrogate time-series.

\section*{References}

\bibliography{mybibfile3}

\begin{thebibliography}{10}
\expandafter\ifx\csname url\endcsname\relax
  \def\url#1{\texttt{#1}}\fi
\expandafter\ifx\csname urlprefix\endcsname\relax\def\urlprefix{URL }\fi
\expandafter\ifx\csname href\endcsname\relax
  \def\href#1#2{#2} \def\path#1{#1}\fi

\bibitem{lacasa2015network}
L.~Lacasa, V.~Nicosia, V.~Latora, Network structure of multivariate time
  series, Scientific reports 5 (2015) 15508.

\bibitem{gao2017complex}
Z.-K. Gao, M.~Small, J.~Kurths, Complex network analysis of time series, EPL
  (Europhysics Letters) 116~(5) (2017) 50001.

\bibitem{dsouza2017exploring}
A.~M. DSouza, A.~Z. Abidin, L.~Leistritz, A.~Wism{\"u}ller, Exploring
  connectivity with large-scale {Granger} causality on resting-state functional
  {MRI}, Journal of Neuroscience Methods 287 (2017) 68--79.

\bibitem{dsouza2018mutual}
A.~M. DSouza, A.~Z. Abidin, U.~Chockanathan, G.~Schifitto, A.~Wism{\"u}ller,
  Mutual connectivity analysis of resting-state functional {MRI} data with
  local models, NeuroImage 178 (2018) 210--223.

\bibitem{Granger1969}
C.~W. Granger, Investigating causal relations by econometric models and
  cross-spectral methods, Econometrica: Journal of the Econometric Society
  (1969) 424--438.

\bibitem{chen2004analyzing}
Y.~Chen, G.~Rangarajan, J.~Feng, M.~Ding, Analyzing multiple nonlinear time
  series with extended {Granger} causality, Physics Letters A 324~(1) (2004)
  26--35.

\bibitem{Stephan2010}
K.~E. Stephan, K.~J. Friston,
  \href{http://www.ncbi.nlm.nih.gov/pubmed/21209846$\backslash$nhttp://www.pubmedcentral.nih.gov/articlerender.fcgi?artid=PMC3013343$\backslash$nhttp://www.pubmedcentral.nih.gov/articlerender.fcgi?artid=3013343{\&}tool=pmcentrez{\&}rendertype=abstract}{{Analyzing
  effective connectivity with fMRI.}}, Wiley interdisciplinary reviews.
  Cognitive science 1~(3) (2010) 446--459.
\newblock \href {http://dx.doi.org/10.1002/wcs.58} {\path{doi:10.1002/wcs.58}}.
\newline\urlprefix\url{http://www.ncbi.nlm.nih.gov/pubmed/21209846$\backslash$nhttp://www.pubmedcentral.nih.gov/articlerender.fcgi?artid=PMC3013343$\backslash$nhttp://www.pubmedcentral.nih.gov/articlerender.fcgi?artid=3013343{\&}tool=pmcentrez{\&}rendertype=abstract}

\bibitem{Blinowska2004}
K.~J. Blinowska, R.~Ku{\'{s}}, M.~Kami{\'{n}}ski,
  \href{http://journals.aps.org/pre/abstract/10.1103/PhysRevE.70.050902}{{Granger
  causality and information flow in multivariate processes.}}, Physical review.
  E, Statistical, nonlinear, and soft matter physics 70~(5 Pt 1) (2004) 050902.
\newblock \href {http://dx.doi.org/10.1103/PhysRevE.70.050902}
  {\path{doi:10.1103/PhysRevE.70.050902}}.
\newline\urlprefix\url{http://journals.aps.org/pre/abstract/10.1103/PhysRevE.70.050902}

\bibitem{Geweke1984}
J.~F. Geweke, \href{http://www.jstor.org/stable/2288723}{{Measures of
  conditional linear dependence and feedback between time series}}, Journal of
  the American Statistical Association 79~(388) (1984) 907--915.
\newblock \href {http://dx.doi.org/10.2307/2288723}
  {\path{doi:10.2307/2288723}}.
\newline\urlprefix\url{http://www.jstor.org/stable/2288723}

\bibitem{angelini2010redundant}
L.~Angelini, M.~De~Tommaso, D.~Marinazzo, L.~Nitti, M.~Pellicoro,
  S.~Stramaglia, Redundant variables and {Granger} causality, Physical Review E
  81~(3) (2010) 037201.

\bibitem{sugihara2012detecting}
G.~Sugihara, R.~May, H.~Ye, C.-h. Hsieh, E.~Deyle, M.~Fogarty, S.~Munch,
  Detecting causality in complex ecosystems, Science 338~(6106) (2012)
  496--500.

\bibitem{bischi2010nonlinear}
G.~I. Bischi, C.~Chiarella, L.~Gardini, Nonlinear Dynamics in Economics,
  Finance and Social Sciences, Springer, 2010.

\bibitem{sugihara1998nonlinear}
G.~Sugihara, R.~M. May, Nonlinear forecasting as a way of distinguishing chaos
  from, Nonlinear Physics for Beginners: Fractals, Chaos, Solitons, Pattern
  Formation, Cellular Automata and Complex Systems (1998) 118.

\bibitem{liao2009kernel}
W.~Liao, D.~Marinazzo, Z.~Pan, Q.~Gong, H.~Chen, Kernel {Granger} causality
  mapping effective connectivity on fmri data, IEEE transactions on medical
  imaging 28~(11) (2009) 1825--1835.

\bibitem{marinazzo2008kernel}
D.~Marinazzo, M.~Pellicoro, S.~Stramaglia, Kernel-{Granger} causality and the
  analysis of dynamical networks, Physical review E 77~(5) (2008) 056215.

\bibitem{marinazzo2011nonlinear}
D.~Marinazzo, W.~Liao, H.~Chen, S.~Stramaglia, Nonlinear connectivity by
  {Granger} causality, Neuroimage 58~(2) (2011) 330--338.

\bibitem{ma2014detecting}
H.~Ma, K.~Aihara, L.~Chen, Detecting causality from nonlinear dynamics with
  short-term time series, Scientific reports 4.

\bibitem{koller2009probabilistic}
D.~Koller, N.~Friedman, F.~Bach, Probabilistic graphical models: principles and
  techniques, MIT press, 2009.

\bibitem{baccala2001partial}
L.~A. Baccal{\'a}, K.~Sameshima, Partial directed coherence: a new concept in
  neural structure determination, Biological cybernetics 84~(6) (2001)
  463--474.

\bibitem{CST2}
A.~Leontitsis, Chaotic systems toolbox,
  https://www.mathworks.com/matlabcentral/fileexchange/1597-chaotic-systems-toolbox
  (2020).

\bibitem{zachary1977information}
W.~W. Zachary, An information flow model for conflict and fission in small
  groups, Journal of anthropological research 33~(4) (1977) 452--473.

\bibitem{thomas2013pathways}
J.~B. Thomas, M.~R. Brier, A.~Z. Snyder, F.~F. Vaida, B.~M. Ances, {Pathways to
  neurodegeneration Effects of HIV and aging on resting-state functional
  connectivity}, Neurology (2013) 10--1212.

\bibitem{abidin2018alteration}
A.~Z. Abidin, A.~M. DSouza, M.~B. Nagarajan, L.~Wang, X.~Qiu, G.~Schifitto,
  A.~Wism{\"u}ller, {Alteration of brain network topology in HIV-associated
  neurocognitive disorder: A novel functional connectivity perspective},
  NeuroImage: Clinical 17 (2018) 768--777.

\bibitem{Ryali2011}
S.~Ryali, K.~Supekar, T.~Chen, V.~Menon, {Multivariate dynamical systems models
  for estimating causal interactions in fMRI}, NeuroImage 54~(2) (2011)
  807--823.
\newblock \href {http://dx.doi.org/10.1016/j.neuroimage.2010.09.052}
  {\path{doi:10.1016/j.neuroimage.2010.09.052}}.

\bibitem{Granger1980}
C.~W.~J. Granger, {Testing for causality. A personal viewpoint}, Journal of
  Economic Dynamics and Control 2~(C) (1980) 329--352.
\newblock \href {http://dx.doi.org/10.1016/0165-1889(80)90069-X}
  {\path{doi:10.1016/0165-1889(80)90069-X}}.

\bibitem{li2010nonlinear}
X.~Li, G.~Marrelec, R.~F. Hess, H.~Benali, A nonlinear identification method to
  study effective connectivity in functional mri, Medical image analysis 14~(1)
  (2010) 30--38.

\bibitem{cao1997practical}
L.~Cao, Practical method for determining the minimum embedding dimension of a
  scalar time series, Physica D: Nonlinear Phenomena 110~(1-2) (1997) 43--50.

\bibitem{mazziotta2001probabilistic}
J.~Mazziotta, A.~Toga, A.~Evans, P.~Fox, J.~Lancaster, K.~Zilles, R.~Woods,
  T.~Paus, G.~Simpson, B.~Pike, et~al., {A probabilistic atlas and reference
  system for the human brain: International Consortium for Brain Mapping
  (ICBM)}, Philosophical Transactions of the Royal Society of London. Series B:
  Biological Sciences 356~(1412) (2001) 1293--1322.

\bibitem{jenkinson2001global}
M.~Jenkinson, S.~Smith, A global optimisation method for robust affine
  registration of brain images, Medical image analysis 5~(2) (2001) 143--156.

\bibitem{sikka2014towards}
S.~Sikka, B.~Cheung, R.~Khanuja, S.~Ghosh, C.~Yan, Q.~Li, J.~Vogelstein,
  R.~Burns, S.~Colcombe, C.~Craddock, et~al., Towards automated analysis of
  connectomes: The configurable pipeline for the analysis of connectomes
  (c-pac), in: 5th INCF Congress of Neuroinformatics, Munich, Germany, Vol.~10,
  2014.

\bibitem{Smith2004}
S.~M. Smith, M.~Jenkinson, M.~W. Woolrich, C.~F. Beckmann, T.~E. Behrens,
  H.~Johansen-Berg, P.~R. Bannister, M.~De~Luca, I.~Drobnjak, D.~E. Flitney,
  et~al., {Advances in functional and structural MR image analysis and
  implementation as FSL.}, NeuroImage 23 Suppl 1 (2004) S208--19.
\newblock \href {http://dx.doi.org/10.1016/j.neuroimage.2004.07.051}
  {\path{doi:10.1016/j.neuroimage.2004.07.051}}.

\bibitem{Wismuller2002}
A.~Wism{\"{u}}ller, O.~Lange, D.~R. Dersch, G.~L. Leinsinger, K.~Hahn,
  B.~P{\"{u}}tz, D.~Auer, {Cluster analysis of biomedical image time-series},
  International Journal of Computer Vision 46~(2) (2002) 103--128.
\newblock \href {http://dx.doi.org/10.1023/A:1013550313321}
  {\path{doi:10.1023/A:1013550313321}}.

\bibitem{tzourio2002automated}
N.~Tzourio-Mazoyer, B.~Landeau, D.~Papathanassiou, F.~Crivello, O.~Etard,
  N.~Delcroix, B.~Mazoyer, M.~Joliot, Automated {anatomical labeling of
  activations in SPM using a macroscopic anatomical parcellation of the MNI MRI
  single-subject brain}, Neuroimage 15~(1) (2002) 273--289.

\bibitem{antinori2007updated}
A.~Antinori, G.~Arendt, J.~Becker, B.~Brew, D.~Byrd, M.~Cherner, D.~Clifford,
  P.~Cinque, L.~Epstein, K.~Goodkin, et~al., Updated research nosology for
  {HIV-associated neurocognitive disorders}, Neurology 69~(18) (2007)
  1789--1799.

\bibitem{kendall1955rank}
M.~G. Kendall, Rank correlation methods (1955).

\bibitem{freund1997decision}
Y.~Freund, R.~E. Schapire, A decision-theoretic generalization of on-line
  learning and an application to boosting, Journal of computer and system
  sciences 55~(1) (1997) 119--139.

\bibitem{hall2009weka}
M.~Hall, E.~Frank, G.~Holmes, B.~Pfahringer, P.~Reutemann, I.~H. Witten, The
  {WEKA} data mining software: an update, ACM SIGKDD explorations newsletter
  11~(1) (2009) 10--18.

\bibitem{takens1981detecting}
F.~Takens, Detecting strange attractors in turbulence, in: Dynamical systems
  and turbulence, Warwick 1980, Springer, 1981, pp. 366--381.

\bibitem{deyle2011generalized}
E.~R. Deyle, G.~Sugihara, Generalized theorems for nonlinear state space
  reconstruction, PLoS One 6~(3) (2011) e18295.

\bibitem{schreiber1996improved}
T.~Schreiber, A.~Schmitz, Improved surrogate data for nonlinearity tests,
  Physical Review Letters 77~(4) (1996) 635.

\end{thebibliography}

\end{document}